\documentclass[10pt,journal,compsoc]{IEEEtran}
\usepackage{ amssymb }
\usepackage{amsmath}
\usepackage{amsthm}
\usepackage{bm}
\usepackage{graphicx}
\usepackage{xcolor}
\newcommand{\bX}{\mathbf{X}}
\newcommand{\bx}{\mathbf{x}}
\newcommand{\bA}{\mathbf{A}}
\newcommand{\bH}{\mathbf{H}}
\newcommand{\bU}{\mathbf{U}}
\newcommand{\bu}{\mathbf{u}}

\newcommand{\br}{\mathbf{r}}

\newcommand{\group}{\mathcal{R}_\mathbf{Q}}

\ifCLASSOPTIONcompsoc
\usepackage[nocompress]{cite}
\else
\usepackage{cite}
\fi

\ifCLASSINFOpdf

\else

\fi

\begin{document}
	
	\title{Group Contrastive Self-Supervised Learning on Graphs}
	
	\author{Xinyi Xu,
		Cheng Deng*,~\IEEEmembership{Senior Member,~IEEE,}
		Yaochen Xie,
		and~Shuiwang~Ji*,~\IEEEmembership{Senior Member,~IEEE}
		\IEEEcompsocitemizethanks{\IEEEcompsocthanksitem X. Xu and C. Deng are with the School of Electronic Engineering, Xidian University, Xi'an 710071, China. E-mail: \{xyxu.xd, chdeng.xd\}@gmail.com.\protect\\
			
			\IEEEcompsocthanksitem Y. Xie and S. Ji are with the Department of Computer Science and Engineering, Texas A\&M University, College Station, TX 77843. E-mail: \{ethanycx, sji\}@tamu.edu}
        \thanks{This work was performed while the first author is visiting Texas A\&M University. *Correspondence should be addressed to these authors.}
		\thanks{Manuscript received xxx}}

	\markboth{Preprint}%
	{Shell \MakeLowercase{\textit{et al.}}: Bare Advanced Demo of IEEEtran.cls for IEEE Computer Society Journals}
	
	\IEEEtitleabstractindextext{
		\begin{abstract}
		We study self-supervised learning on graphs using contrastive methods. A general scheme of prior methods is to optimize two-view representations of input graphs. In many studies, a single graph-level representation is computed as one of the contrastive objectives, capturing limited characteristics of graphs. We argue that contrasting graphs in multiple subspaces enables graph encoders to capture more abundant characteristics. To this end, we propose a group contrastive learning framework in this work. Our framework embeds the given graph into multiple subspaces, of which each representation is prompted to encode specific characteristics of graphs. To learn diverse and informative representations, we develop principled objectives that enable us to capture the relations among both intra-space and inter-space representations in groups. Under the proposed framework, we further develop an attention-based representor function to compute representations that capture different substructures of a given graph. Built upon our framework, we extend two current methods into GroupCL and GroupIG, equipped with the proposed objective. Comprehensive experimental results show our framework achieves a promising boost in performance on a variety of datasets. In addition, our qualitative results show that features generated from our representor successfully capture various specific characteristics of graphs.
		\end{abstract}
		
	}
	
	\maketitle
	\IEEEdisplaynontitleabstractindextext
	
	\IEEEpeerreviewmaketitle

	\ifCLASSOPTIONcompsoc
	\IEEEraisesectionheading{\section{Introduction}\label{sec:introduction}}
	\else
	\section{Introduction}
	\label{sec:introduction}
	\fi
    With the advances of deep learning, graph neural networks~\cite{kipf2016semi, gao2019graph, velivckovic2017graph, liu2020deep} (GNNs) have been developed for learning from graph-structured data such as molecules and proteins. GNNs have achieved great success on various graph learning tasks~\cite{velivckovic2017graph, gao2019graph, gao2018large, yuan2020structpool, rong2020self, wang2020second}. However, such success hinges on a large amount of labeled data, which is expensive and even not available. To mitigate the dependence on labels, self-supervised learning (SSL)~\cite{xie2020noise2same, wang2021contrastive, you2020graph, chen2020simple, pathak2016context, doersch2015unsupervised} is proposed to use supervisions from graphs themselves. SSL is initially investigated for unsupervised image tasks~\cite{chen2020simple, pathak2016context, doersch2015unsupervised, xie2020noise2same} and later successfully applied on unsupervised sequence tasks~\cite{devlin2018bert, wang2019self, yang2019xlnet}. Inspired by its success in both image and sequence domains, a variety of SSL methods based on GNNs are proposed~\cite{you2020graph, sun2019infograph, thakoor2021bootstrapped, zhu2020deep, jiao2020sub, peng2020graph, velivckovic2018deep, hamilton2020graph, wang2017mgae, kipf2016variational}. Among them, contrastive learning (CL) methods~\cite{you2020graph, sun2019infograph, thakoor2021bootstrapped, zhu2020deep, jiao2020sub, peng2020graph, velivckovic2018deep} are becoming the mainstream approaches in this field. CL methods train models on pretext tasks that encode the agreement between two views of representations. These two views can be global-local pairs~\cite{sun2019infograph, hassani2020contrastive, tschannen2019mutual,hjelm2018learning} or differently transformed graph data~\cite{you2020graph, hassani2020contrastive, wang2021contrastive, rong2020self}. The learning goal is to make these two-view representations similar if they are from the same graph and dissimilar if they are from different graphs.
    
    Existing CL methods usually compute graph representations from a single perspective as components of the contrastive objectives~\cite{you2020graph, sun2019infograph, zhu2020deep, thakoor2021bootstrapped}. However, we argue that contrasting two graphs in multiple subspaces has the potential of capturing more abundant characteristics of graphs. In domains of proteins and images, studying data from multiple perspectives has been shown to effectively capture powerful features. For example, previous studies~\cite{shechtman2007matching, lampert2013attribute} extract the feature of images by investigating the similarities of various local parts progressively. Other methods~\cite{yu2017protein, kim2010walk} study the properties of protein sequences by exploring diverse gene sub-sequences. Based on this motivation, we propose a group contrastive learning framework for graphs in this work. Different from existing methods that perform contrastive learning in a single space, our framework embeds a given graph into representations in various subspaces and perform contrastive learning in each subspaces. For simplicity, we refer to a group as a set of representations of different graph views within the same subspace. Note that when the number of groups is set to one, our framework reduces to prior methods without group contrast. 
    
    We investigate the agreement for each group which encourages the representation to encode one specific characteristic of the given graph. We design our objective based on the mutual information (MI) to capture both intra-space and inter-space relations. More specifically, we propose to maximize the MI between two views of representations in the same group while minimizing the MI between the representations of one view across different groups. To enable the optimization, we derive the MI lower bound under both parametric and nonparametric cases. Under the proposed framework, we further develop an attention-based representor function to compute multiple representations, of which each one is encouraged to focus on some specific nodes, thereby encoding one specific substructure. Note that the idea of using groups has been shown to be effective in the image domain~\cite{chen2019hybrid, opitz2018deep, kim2018attention, xu2020towards}, and here we successfully employ it to the graph SSL task.
    
    Built upon our framework, we extend two previous methods into GroupCL and GroupIG. We evaluate the effectiveness of these two methods on both unsupervised graph classification and transfer learning tasks. Comprehensive quantitative experimental results demonstrate that our methods achieve new state-of-the-art performance on a majority of datasets when compared with previous methods. Particularly, GroupCL and GroupIG show superior performance consistently comparing with the corresponding non-grouping methods across different datasets and tasks. Furthermore, we conduct a visualization experiment using the attention weights learned by our representor function. The qualitative results intuitively illustrate the substructures captured by various representations.
    
	\section{Background and Related Work}
	\subsection{Notations and Problem Setup}
	Let $G = (V, E)$ denote a graph with the node set $V = \{v_{1}, v_{2}, \cdots, v_{N}\}$ and the edge set $E \subseteq V \times V$, where $N$ denotes the number of nodes in this graph. The node set is initially represented by a node feature matrix $\bX \in \mathbb{R}^{N\times d}$, where $d$ denotes the feature dimension. And the edge set is represented by an adjacency matrix $\bA \in \mathbb{R}^{N \times N}$, of which the element is determined by $\bA_{ij} = \mathbf{1}[(v_{i}, v_{j}) \in E]$. We are interested in the unsupervised graph representation learning task. Given graph data $(\bA, \bX)$, the goal is to learn a graph-level encoder, $\mathcal{E}: (\mathbb{R}^{N \times d}, \mathbb{R}^{N \times N}) \rightarrow \mathbb{R}^{1 \times d_{o}}$ to encode the graph into a high-level representation of dimension $d_{o}$, formulated as $\mathbf{h} = \mathcal{E}(\bX, \bA)\in \mathbb{R}^{1 \times d_{o}}$. In particular, a graph-level encoder usually consists of a node encoder $\mathcal{E}_{n}: (\mathbb{R}^{N \times d}, \mathbb{R}^{N \times N}) \rightarrow \mathbb{R}^{N \times d_{n}}$ and a readout function $\mathcal{R}:\mathbb{R}^{N \times d_{n}} \to \mathbb{R}^{1 \times d_{o}}$, where $d_n$ denotes the dimension of node embeddings.
	The node encoder computes the node embedding matrix from $(\bA, \bX)$, \textit{i.e.}, $\bH = \mathcal{E}_{n}(\bX, \bA)$. And the readout function summarize the node embeddings into the desired graph-level embedding, \textit{i.e.}, $\mathbf{h}=\mathcal{R}(\bH)$.

	\subsection{Graph Neural Networks}
    Graph neural networks (GNNs)~\cite{xu2018powerful, kipf2016semi, velivckovic2017graph} have demonstrated their effectiveness in learning the representation of graph-structured data such as molecules and social networks. GNNs iteratively update the representation of each node by aggregating information from their neighbor nodes, aiming at capturing the local structural information. For each node, a single GNN-layer aggregates information from its $1$-hop neighborhood. Stacking $L$ aggregation layers hence enables each node representation to capture information within the $L$-hop neighborhood.
	To be concrete, the update of the $\ell$-th layer in an $L$-layer GNN can be described as
	\begin{equation}
	\begin{split}
	\mathbf{a}_{v}^{(\ell)} &= \text{AGGREGATE}^{(\ell)}\Big(\{\mathbf{h}_{v^{\prime}}^{(\ell-1)}: v^{\prime} \in (\mathcal{N}(v) \cup v) \}\Big), \\
	\mathbf{h}_{v}^{(\ell)} &= \text{COMBINE}^{(\ell)}\Big(\mathbf{a}_{v}^{(\ell)}, \mathbf{h}_{v}^{(\ell-1)}\Big),
	\end{split}
	\end{equation}
	where $\mathbf{h}_{v}^{(\ell)}$ denotes the feature vector of node $v$ at the $\ell$-th layer, the initial $\mathbf{h}_{v}^{(0)}$ is set to input node features, $\bx_{v}$. $\mathbf{h}_{v}^{(L)}$ denotes the ultimate representation of node $v$, and $\mathcal{N}(v)$ is a set of vertices that connect to node $v$.
	There are different types of $\text{AGGREGATE}^{(\ell)}(\cdot)$ and $\text{COMBINE}^{(\ell)}(\cdot)$ functions. For instance, GIN~\cite{xu2018powerful} integrates the $\text{AGGREGATE}^{(\ell)}(\cdot)$ and $\text{COMBINE}^{(\ell)}(\cdot)$ functions as:
	\begin{equation}
	\mathbf{h}_{v}^{(\ell)} = \text{MLP}^{(\ell)} \Big( \big(1 + \epsilon^{(\ell)} \big) * \mathbf{h}_{v}^{(\ell-1)} + \sum_{v^{\prime} \in \mathcal{N}(v)}, \mathbf{h}_{v^{\prime}}^{(\ell-1)} \Big),
	\end{equation}
	where $\epsilon^{(\ell)}$ can be a learnable parameter or a fixed scalar.
	For the graph-level task, a READOUT function is employed to summarize the ultimate node embeddings, then an optional projection head, usually a multi-layer preceptron (MLP), is used to perform linear projection to generate the graph-level embedding. Mathematically,
	\begin{equation}
	\mathbf{h} = \text{MLP}\Big( \text{READOUT} \big( \{\mathbf{h}_{v}^{(L)}: v \in V \} \big) \Big)
	\end{equation}

	\subsection{Graph Contrastive Learning}
	We take advantage of the contrastive learning (CL) technique to solve the unsupervised learning issue. We describe the graph contrastive learning framework in a two-view case. The contrastive learning is performed between the representations of two views $u$ and $r$. It aims to enlarge the agreement between representations of the positive pairs, \textit{i.e.}, two views associated with the same graph instance, and weakening that of the negative pairs, \textit{i.e.}, two views associated with different instances. Views $u$ and $r$ of a graph are usually generated by data transformation functions denoted by $\mathcal{T}_{u}$ and $\mathcal{T}_{r}$. Previous studies implement $\mathcal{T}_{u}$ and $\mathcal{T}_{r}$ in diverse manners. We specifically introduce two approaches to generate views in Sections~\ref{sec: Group GraphCL} and~\ref{sec: Group InfoGraph} respectively.

	\section{Group Contrastive Learning on Graphs}
	In self-supervised learning, the contrastive learning technique has demonstrated its capability in learning representations without labels. Some previous studies perform the contrastive process between the representations of two augmented views or the global-and-local pair. However, they share a common drawback. That is, they contrast two objectives in a single space, which captures limited characteristics. 
	For images and proteins, previous approaches study the objectives through contrast various local parts and sub-sequences respectively. Thus, we argue that contrasting two graphs in multiple subspaces has the potential of capturing more abundant characteristics of graphs. To this end, we propose a graph group contrastive learning framework.
	Overall, our framework embeds the given graph into various subspaces, resulting in multiple groups of representations for two views. We investigate the agreement of representations group-wisely which aims at enabling these representations to encode various characteristics of the given graph.

	\subsection{The Proposed Group Contrastive Learning Framework} \label{sec_method}
    Our proposed framework considers two views $u$ and $r$ as the main view and auxiliary view, respectively, in which case the processing procedures may be asymmetric. 
    For each branch of view, our framework performs data transformations on the given graph to obtain the view and employ an individual GNN-based encoder to compute the corresponding multiple representations. In particular, the main branch computes multiple representations based on different learnable parameters, formulated as
	\begin{equation}
        [\mathbf{u}^{(1)}, \mathbf{u}^{(2)}, \dots, \mathbf{u}^{(p)}] = \mathcal{E}_{m}(\mathcal{T}_{u}(\bX, \bA)),\\
	\end{equation}
	where $p$ denotes the number of representations and $\bu^{(i)} \in \mathbb{R}^{d_{o}/p}$ is the $i$-th graph-level embedding. Here, $\mathcal{E}_{m}$ is a multi-representation encoder which seeks to encode multiple characteristics of the given graph.
	In contrast, there are two approaches to compute the multiple representations for the auxiliary view $r$. One can employ an encoder that only computes one representation, and then duplicates it for $p$ times to obtain $p$ feature vectors, given by
	\begin{equation}
    	\mathbf{r} = \mathcal{E}_{r}(\mathcal{T}_{r}(\bX, \bA)),
	\end{equation}
	\begin{equation}
    	[\mathbf{r}^{(1)}, \ \mathbf{r}^{(2)}, \dots, \ \mathbf{r}^{(p)}] = \text{Duplicate}_p(\br),
	\end{equation}
	where $\mathcal{E}_{r}$ is the encoder of view $r$. In an alternative way, we can use the same encoder as branch $u$ and produce multiple feature vectors, given by
	\begin{equation}
    	[\mathbf{r}^{(1)}, \ \mathbf{r}^{(2)}, \dots, \ \mathbf{r}^{(p)}] = \mathcal{E}_{m}(\mathcal{T}_{r}(\bX, \bA)).
	\end{equation}
	Given representations of those two views, We formulate $p$ groups of representations, those are $\{(\bu^{(i)}, \br^{(i)}) |i=1, 2, \cdots, p\}$. we perform $p$ times contrastive learning studies between representations of those groups.
	
	During prediction, we input one graph into the learned multi-representation encoder $\mathcal{E}_{m}$ for the main branch without performing transformations, resulting in $p$ feature vectors. We then combine these $p$ vectors to jointly represent the graph.
	Here, we adopt the simple concatenation operation. Formally,
	\begin{equation}
    	[\mathbf{h}^{(1)}, \ \mathbf{h}^{(2)}, \dots, \ \mathbf{h}^{(p)}] = \mathcal{E}_{m}(\bX, \bA),
	\end{equation}
	\begin{equation}
	\mathbf{h} = \text{Concat}([\mathbf{h}^{(1)}, \mathbf{h}^{(2)}, \dots, \mathbf{h}^{(p)}]).
	\end{equation}
	Here $\mathbf{h}$ is a vector of size $d_{o}$ which is computed by $p \times (d_{o}/p)$.  
	
	\subsection{Intra-Space Objective Function}
    As introduced in Section~\ref{sec_method}, the output of our group contrastive learning framework is $p$ groups of representation regarding two views, \textit{i.e.}, 
    $\{(\bu^{(k)}, \br^{(k)})|k=1, 2, \cdots, p\}$. Our goal is to optimize $\mathcal{E}_{m}$ such that the vectors in $\bU$ encode diverse characteristics of the input graph. To achieve this goal, we employ two essential objectives based on mutual information (MI) regarding both intra-space representation pairs and inter-space representation pairs, respectively. 

    For the intra-space objective, we seek to maximize the mutual information between representations of two views within each group. The intra-space MI maximization enables the learning of informative representation for each individual group. In particular, the paired representations are $\{(\bu^{(1)}, \br^{(1)}), (\bu^{(2)}, \br^{(2)}), \cdots, (\bu^{(p)}, \br^{(p)})\}$. We hence formulate the intra-space objective as

	\begin{equation}
	\max_{\theta, \phi} \sum_{k=1}^{p} \textbf{MI}\big(\bu^{(k)}, \br^{(k)}\big),
	\label{eq: intra-space}
	\end{equation}
	where $\theta, \phi$ are parameters of encoders in the main-view branch and the auxiliary-view branch, respectively. As the mutual information $\textbf{MI}(\bu^{(k)}, \br^{(k)})$ becomes intractable when distributions of $\br^{(k)}$ and $\bu^{(k)}$ are unknown, a common substitute approach to maximizing MI is to maximize its lower-bound estimation based on the sampled examples~\cite{xie2021self, nowozin2016f}. Among all existing MI lower bounds, we adopt the Jensen-Shannon estimator of MI, which is computed as
	\begin{equation}
	\begin{split}
	\textbf{MI}_{JS}(\bu^{(k)}, \br^{(k)}) := &\mathbb{E}_{P(\bu^{(k)}, \br^{(k)})}\Big[
	-\text{SP}\Big(
	-\mathcal{D}\big(\bu^{(k)}, \br^{(k)}\big)
	\Big)\Big] -  \\  &\mathbb{E}_{P(\bu^{(k)})}\mathbb{E}_{P(\br^{(k)})}\Big[
	\text{SP}\Big(
	\mathcal{D}\big(\bu^{(k)}, \br^{(k)}\big)
	\Big)\Big],
	\end{split} \label{eq: JS}
	\end{equation}
	where $\text{SP}(x) = \log(1 + e^{x})$ and $\mathcal{D(\cdot,\cdot)}: \mathbb{R}^{d} \times \mathbb{R}^{d} \rightarrow \mathbb{R}$ is a discriminator that takes  representations of two views as inputs, and scores the agreement between them. We simply instantiate the discriminator as the dot product between two representations, \textit{i.e.}, $\mathcal{D}(\bu^{(k)}_{i}, \br^{(k)}_{i}) = \big(\bu^{(k)}_{i}\big)^T \br^{(k)}_{i}.$
    
	\subsection{Inter-Space Objective Function}
	\label{sec: inter-group}
	In addition to the intra-space objective, we also constraint the pairwise relation across different groups of the same view to enforce the diversity of inter-space. In particular, we propose to employ an inter-space optimization objective based on mutual information minimization. The objective prompts any two representations in the same view to capture different characteristics of the given graph. The inter-space objective focuses on each pair of representations across different groups within view $u$, \textit{i.e.}, $\{(\bu^{(k)}, \bu^{(l)})|1\le k<l\le p\}$, and can be formulated as
    
	\begin{equation}
	\min_{\theta} \frac{1}{2}\sum_{k=1}^{p}\sum_{l=1}^{p} \textbf{MI}\big(\bu^{(k)}, \bu^{(l)}\big).
	\label{mi2}
	\end{equation}
	In order to minimize the MI, we introduce an upper bound of MI as an efficient estimation. The upper bound MI we adopt here is based on the contrastive log-ratio upper bound~\cite{cheng2020club} (CLUB). Concretely, for two random variables $x$ and $y$, the upper bound of MI is defined as
	\begin{equation}
	    I_{\text{CLUB}} := \mathbb{E}_{P(x,y)}[\log  P(y|x)] - \mathbb{E}_{P(x)}\mathbb{E}_{P(y)}[\log P(y|x)].
	    \label{eq: club}
	\end{equation}
	To enable the computation of CLUB, a key challenge is to model the intractable conditional distribution $P(y|x)$. We propose two approaches to model $P(y|x)$, \textit{i.e.}, the non-parameterized estimation and the parameterized estimation, based on whether the same dimensions of $x$ and $y$ correspond to each other across different representations.
	For both cases, we follow~\cite{cheng2020club} to assume that the distribution $y$ conditional on $x$ is subject to a Gaussian distribution, of which the mean and variance are determined respectively depending on their concrete situations. 
	
	\textbf{Non-parameterized estimation}. We first consider the case where the correspondence exists between each dimension of $x$ and $y$, two random vectors as different representations of a given graph. In other words, the same dimensions of the two vectors $x_i$ and $y_i$ are associated with the same specific feature. In this case, we introduce the assumption $\mathbb{E}[y|x]=x$ to simplify the computation, \textit{i.e.}, the expectation of $y$ conditional on $x$ equals to $x$. Such an assumption commonly exists in many scenarios such as noise models, where $y$ is considered as observed values and $x$ is considered as their corresponding signals.
	We further assume that the variance $\Sigma$ is a diagonal matrix with the same values on its diagonal, \textit{i.e.}, each dimension of $y$ only depends on the corresponding dimension of $x$ and has equal variance. Concretely, for two representations $u^{(k)}$ and $u^{(l)}$, the distribution of $u^{(l)}$ conditional on $u^{(k)}$ is subject to
	\begin{equation}
	\begin{split}
	P\big(\bu^{(l)}\ \big|\ \bu^{(k)}, \Sigma\big) &= \mathcal{N}\big( \bu^{(l)}
	\ \big|\ 
	\bu^{(k)}, \beta^{-1}\bm I \big)\\
	&= \prod_i \mathcal{N}\big(u^{(l)}_i
	\ \big|\ 
	u^{(k)}_i, \beta^{-1}\big),
	\end{split}
	\label{gau}
	\end{equation}
	where $\beta^{-1}$ denotes the variance shared by all dimensions, $u^{(l)}_i$ and $u^{(k)}_i$ denotes the $i$-th dimension of $\bu^{(l)}$ and $\bu^{(k)}$, respecticely. In Section~4, we further empirically demonstrate the effectiveness of models under the above assumptions.
	Given the assumptions, we are able to simplify the CLUB objective for a more efficient and specific computation of MI upper bound.
	We first rewrite Equation~\eqref{eq: club} into
	\begin{equation}
	\begin{split}
	\hat I_{\text{CLUB}} = &
	\mathbb{E}_{P(\bu^{(k)}, \bu^{(l)})} \log \prod_i \mathcal{N}\big(
	u^{(l)}_{i}
	\ \big|\ 
	u^{(k)}_{i}, \beta^{-1}
	\big) - \\
	& \mathbb{E}_{P(\bu^{(k)})}\mathbb{E}_{P(\bu^{(l)})} \log \prod_i \mathcal{N}\big(
	u^{(l)}_{i}
	\ \big|\ 
	u^{(k)}_{i}, \beta^{-1}\big).\\
	\end{split}\label{eq: club1}
	\end{equation}
	Next, we apply $L_2$-normalization to each representation and rewrite Equation~\eqref{eq: club1} with further simplification, formulated as
	\begin{equation}
	\begin{split}
	&\log \prod_i \mathcal{N}(u^{(l)}_{i}
	\ \big|\ 
	u^{(k)}_{i}, \beta^{-1}) \\
	= & -\frac{\beta}{2}\sum_i\left(u^{(k)}_{i} - u^{(l)}_{i} \right)^2 + \frac{d^{\prime}}{2}\ln \beta - \frac{d^{\prime}}{2}\ln(2\pi) \\
	= &-\frac{\beta}{2} \left\| \bu^{(k)} - \bu^{(l)} \right\|^2 + \frac{d^{\prime}}{2}\ln \beta - \frac{d^{\prime}}{2}\ln(2\pi) \\
	= &-\frac{\beta}{2}\left( \|\bu^{(k)}\|^2 + \|\bu^{(l)}\|^2 - 2\big(\bu^{(k)}\big)^{T}\bu^{(l)} \right)^2 + c \\
	= & \beta\left( \big(\bu^{(k)}\big)^{T}\bu^{(l)}-1\right)+ c.
	\end{split}
	\end{equation}
	where $c = \frac{d^{\prime}}{2}\ln \beta - \frac{d^{\prime}}{2}\ln(2\pi)$ and $d^{\prime}$ denotes the dimension of every representation. 
	Based on this formulation, we can develop Equation~\eqref{eq: club1} as
	\begin{equation}
	    \begin{split}
	        \hat I_{\text{CLUB}} = 
	        & \mathbb{E}_{P(\bu^{(k)}, \bu^{(l)})} \Bigg [\beta\left( \big(\bu^{(k)}\big)^{T}\bu^{(l)}-1\right)+ c\Bigg] + \\
	        & \mathbb{E}_{P(\bu^{(k)})}\mathbb{E}_{P(\bu^{(l)})} \Bigg[\beta\left(1- \big(\bu^{(k)}\big)^{T}\bu^{(l)}\right)+ c\Bigg] \\ \leq
	        & \mathbb{E}_{P(\bu^{(k)}, \bu^{(l)})} \Bigg[\beta\left( \big(\bu^{(k)}\big)^{T}\bu^{(l)}-1\right)+ c\Bigg] + \\ &\mathbb{E}_{P(\bu^{(k)})}\mathbb{E}_{P(\bu^{(l)})} \Big[2\beta +c\Big],
	    \end{split}\label{eq: CLUB}
	\end{equation}
	where the inequality is derived based on $\big(\bu^{(k)}\big)^{T}\bu^{(l)} \geq -1$. Our goal minimizing $I_{\text{CLUB}}$ is hence equivalent to minimizing $ \mathbb{E}_{P(\bu^{(k)}, \bu^{(l)})} \Big[ \big(\bu^{(k)}\big)^{T}\bu^{(l)} \Big]$. To be consistent with Equation ~\eqref{eq: JS}, we also employ the soft plus (SP) function and obtain the final objective
	\begin{equation}
	    \min _{\theta, \phi} \mathbb{E}_{P(\bu^{(k)}, \bu^{(l)})}[\text{SP}(\mathcal{D}(\bu^{(k)}, \bu^{(l)}))].
	\end{equation}
	Note that in Equation~\eqref{eq: CLUB}, we apply a constant upper bound to the term $\mathbb{E}_{P(\bu^{(k)})}\mathbb{E}_{P(\bu^{(l)})}\Big[\beta\left(1- \big(\bu^{(k)}\big)^{T}\bu^{(l)}\right)+ c\Big]$. This is because including this term potentially results in an reversal effect to the intra-space objective in Equation~\eqref{eq: JS}. More specifically, Equation~\eqref{eq: JS} enlarges the agreement between $\bu^{(k)}$ and $\br^{(k)}$, while Equation~\eqref{eq: CLUB} enlarges that between $\bu^{(k)}$ and $\bu^{(l)}$, under the joint distribution. These two goals encourage the distribution of $\bu^{(k)}$ to be similar to two different distributions, being mutually incompatible. 
	
	\textbf{Parameterized estimation.} We then consider the second case where there is no correspondence between dimensions of $x$ and $y$. In this case, we are required to estimate the mean and variance of the conditional Gaussian distribution via learning a parameterized variational distribution~\cite{cheng2020club}.
	Concretely, we employ two independent multi-layer perceptrons (MLP) to generate the mean and variance respectively. Then Equation~\eqref{gau} can be rewritten as  
	\begin{equation}
	\begin{split}
	    P_{\psi, \eta}(\bu^{(l)} | \bu^{(k)}, \Sigma) &= \mathcal{N}\Big(\bu^{(l)} 
	    \ \big| \ 
	    \mu_{\psi}(\bu^{(k)}), 
	    \sigma^{2}_{\eta}(\bu^{(k)}) \Big) \\
	    & = \prod_{i} \mathcal{N}\Bigg( u_{i}^{(l)} 
	    \ \Big| \ 
	    \Big(\mu_{\psi}(\bu^{(k)})\Big)_{i}, 
	    \Big(\sigma^{2}_{\eta}(\bu^{(k)})\Big)_{i} \Bigg)
	\end{split} 
	\label{eq1}
	\end{equation}
	where $\mu_{\psi}(\cdot)$ and $\sigma_{\eta}(\cdot)$ are two MLPs for generating the mean and the standard deviation respectively. $\psi$ and $\eta$ are the parameters of $\mu_{\psi}(\cdot)$ and $\sigma_{\eta}(\cdot)$ respectively. To be simplified, we denote $\mu_{\psi}(\bu^{(k)})$ as $\bm \mu_{\psi}^{(k)}$ and $ \sigma_{\eta}(\bu^{(k)})$ as $\bm \sigma_{\eta}^{(k)}$.
	We rewrite Equation~\eqref{eq1} in its log-form as
	\begin{equation}
	\begin{split}
	    &\log \prod_{i} \mathcal{N}\Big(u_{i}^{(l)}
	    \ \big|\  \mu_{\psi, i}^{(k)}, 
	    \sigma_{\eta, i}^{(k)}
	    \Big) \\
	    =&\sum_{i} \Bigg( 
	    -\frac{1}{2}  
	    \sigma_{\eta, i}^{(k)} 
	    - \frac
	    { \Big(
	    u_{i}^{(l)} -
	    \mu_{\psi, i}^{(k)}
	    \Big)^2} 
	    {2  
	    \sigma_{\eta, i}^{(k)}
	    } + c^{\prime} 
	    \Bigg),
	\end{split}
	\end{equation}
	where $u_{i}^{(l)}$, $\mu_{\psi, i}^{(k)}$, and $\sigma_{\eta, i}^{(k)}$ are the $i$-th dimension of $\bu^{(l)}$, $\bm \mu^{(k)}_{\psi}$, and $\bm \sigma^{(k)}_{\eta}$ respectively. Furthermore, $c^{\prime}=-\frac{1}{2}\log \pi$.
	Based on this formulation, we can further derive the CLUB as following
	\begin{equation}
	    \begin{split}
	       &\hat I_{\text{CLUB-para}} \\
	       := &
    	\mathbb{E}_{P(\bu^{(k)}, \bu^{(l)})}
    	\Bigg[\sum_{i}\Bigg( 
	    -\frac{1}{2}  
	    \sigma_{\eta, i}^{(k)} 
	    - \frac
	    { \Big(u_{i}^{(l)} - 
	    \mu_{\psi, i}^{(k)}
	    \Big)^2} 
	    {2 
	    \sigma_{\eta, i}^{(k)}
	    }+c^{\prime} \Bigg) \Bigg]+ \\
    	& \mathbb{E}_{P(\bu^{(k)})}\mathbb{E}_{P(\bu^{(l)})} \Bigg[\sum_{i} \Bigg( 
	    \frac{1}{2}  
	    \sigma_{\eta, i}^{(k)} 
	    + \frac
	    { \Big(u_{i}^{(l)} - 
	    \mu_{\psi, i}^{(k)}
	    \Big)^2} {2 
	    \sigma_{\eta, i}^{(k)}
	    }+c^{\prime} \Bigg)\Bigg]. 
	    \end{split}
	\end{equation}
	We remove the second term in this equation in the same manner as used in Equation~\eqref{eq: CLUB}, so the MI minimization problem is equivalent to 
	\begin{equation}
	\begin{split}
	    &\min \textbf{MI}(\bu^{(k)}, \bu^{(l)}) \\
	    := &\min_{\theta, \phi}
    	\mathbb{E}_{P(\bu^{(k)}, \bu^{(l)})} \Bigg[\sum_{i}\Bigg(-
	    \sigma_{\eta, i}^{(k)}
	    - \frac{\Big(u_{i}^{(l)} -
	    \mu_{\psi, i}^{(k)}
	    \Big)^2} {
	    \sigma_{\eta, i}^{(k)}
	    }\Bigg)\Bigg]
	\end{split}
	\end{equation}
	
	To determine the parameters used to generate the mean and variance, \textit{i.e.} $\psi$ and $\eta$, we maximize the likelihood by
	\begin{equation}
	    \max_{\psi} \mathbb{E}_{P(\bu^{(k)}, \bu^{(l)})}
	    \Bigg[\sum_{i} \Bigg(-
	    \sigma_{\eta, i}^{(k)}
	    - \frac
	    { \Big(u_{i}^{(l)} - 
	    \mu_{\psi, i}^{(k)}
	    \Big)^2} 
	    { 
	    \sigma_{\eta, i}^{(k)}
	    } \Bigg)\Bigg]
	\end{equation}
	To sum up, the parameters involved in the conditional distribution and the encoder are trained adversarially by 
	\begin{equation}
	    \max_{\psi, \eta} \min_{\theta, \phi}
    	\mathbb{E}_{P(\bu^{(k)}, \bu^{(l)})} \Bigg[\sum_{i}\Bigg( 
	    -
	    \sigma_{\eta, i}^{(k)}
	    - \frac
	    { \Big(u_{i}^{(l)} - 
	    \mu_{\psi, i}^{(k)}
	    \Big)^2}
	    { 
	    \sigma_{\eta, i}^{(k)}
	    }\Bigg)\Bigg]
	\end{equation}

    \begin{figure*}[!ht]
		\includegraphics[width=0.95\linewidth]{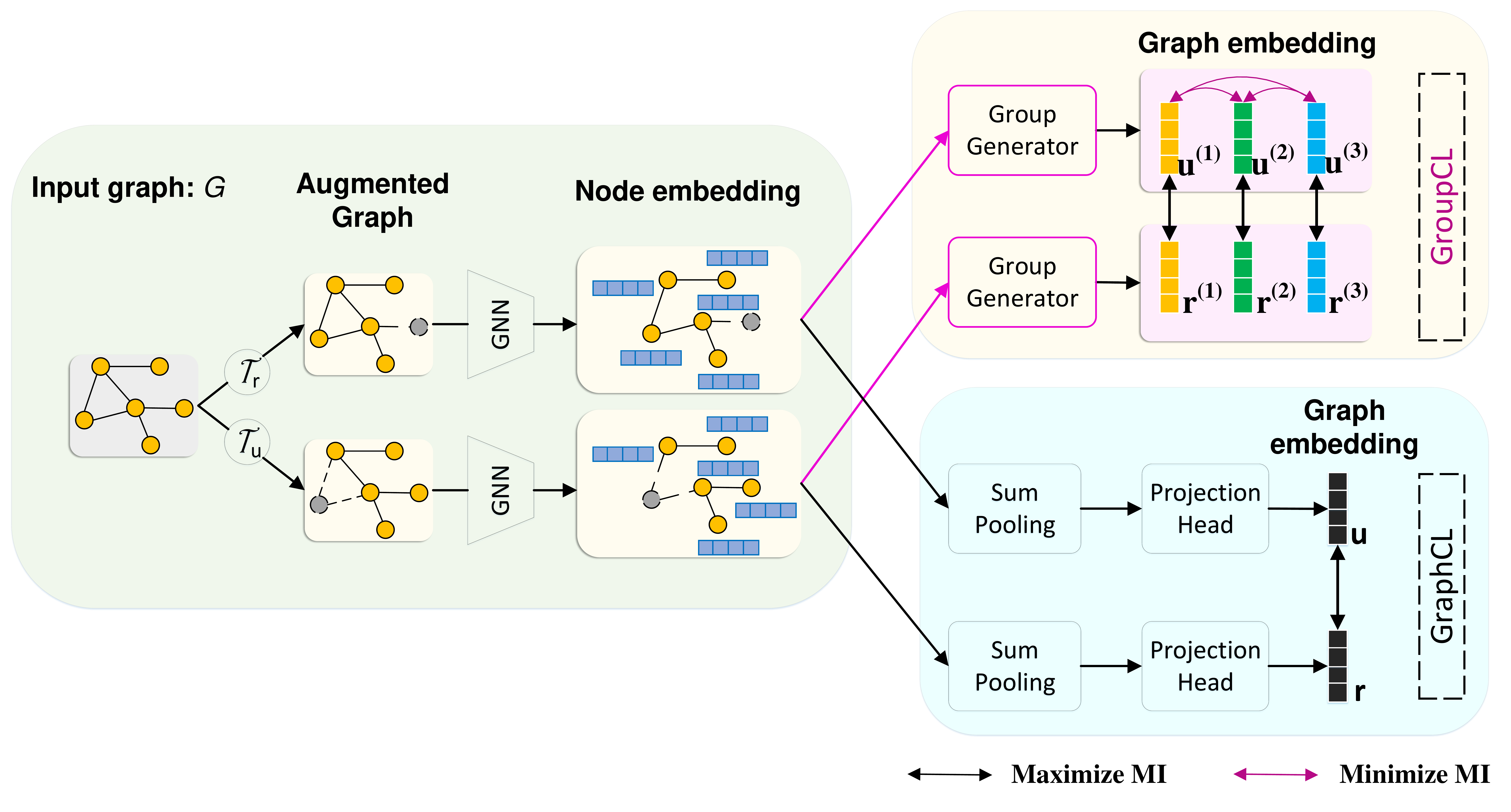}
		\caption{The pipelines of GraphCL and GroupCL. These two methods share the same trunk, in which the data are successively processed by the data augmentation and the GNN encoder. GroupCL employs a representor function $\mathcal{R}_\mathbf{Q}$ to generate multiple graph-level embeddings for two views. The maximizing MI objectives are employed on the representations in the same group from different views. The minimizing MI objectives are employed on the representations in different groups from the same view. GraphCL adopts a sum pooling and projection head to generate one graph embedding.}
		\label{fig:framework}
	\end{figure*}
	
	\subsection{The Overall Objective Function}
	We combine the intra-space and the inter-space objectives and obtain the final objective as 
	\begin{equation} \label{eq: obj}
	\begin{split}
	\max_{\theta, \phi} 
	\sum_{k=1}^{p} \textbf{MI}(\br^{(k)}, \bu^{(k)}) - 
	\lambda * \sum_{k=1}^{p}\sum_{l=k+1}^{p} 
	\textbf{MI}(\bu^{(k)}, \bu^{(l)}),
	\end{split}
	\end{equation}
	which is approximated by the following objective based on the corresponding bounds. We can either adopt the non-parameterized upper bound to form the final objective as
	\begin{equation}\label{eq: final_obj}
	\begin{split}
	\min_{\theta, \phi}\  
	&
	\mathbb{E}_{P(k)}
	\mathbb{E}_{P(\br^{(k)}, \bu^{(k)})} \bigg[\text{SP}\Big(
	-\mathcal{D}(\br^{(k)}, \bu^{(k)})
	\Big)\bigg] +  \\  
	&\mathbb{E}_{P(k)}
	\mathbb{E}_{P(\br^{(k)})}\mathbb{E}_{P(\bu^{(k)})}
	\bigg[\text{SP}\Big(
	\mathcal{D}(\br^{(k)}, \bu^{(k)})
	\Big)\bigg] + \\
	& \lambda * \mathbb{E}_{P(k)} \mathbb{E}_{P(l)} \mathbb{E}_{P(\bu^{(k)}, \bu^{(l)})} \bigg[\text{SP}\Big(
	\mathcal{D}(\bu^{(k)}, \bu^{(l)})
	\Big)\bigg] ,
	\end{split}
	\end{equation}
	or the parameterized bound to form the final objective as
	\begin{equation}\label{eq: final_obj1}
	\begin{split}
	\max_{\psi, \eta} \min_{\theta, \phi}\   
	&\mathbb{E}_{P(k)}
	\mathbb{E}_{P(\br^{(k)}, \bu^{(k)})}
	\bigg[\text{SP}\Big(-\mathcal{D}(\br^{(k)}, \bu^{(k)})\Big)\bigg] +  \\  
	&\mathbb{E}_{P(k)}
	\mathbb{E}_{P(\br^{(k)})}\mathbb{E}_{P(\bu^{(k)})}
	\bigg[\text{SP}\Big(\mathcal{D}(\br^{(k)}, \bu^{(k)})\Big)\bigg] + \\
	& \lambda * \mathbb{E}_{P(k)} \mathbb{E}_{P(l)}
    	\mathbb{E}_{P(\bu^{(k)}, \bu^{(l)})} \bigg[\\
    	&\sum_{i}\bigg(-
	    \sigma_{\eta, i}^{(k)}
	    - \frac{\Big(u_{i}^{(l)} - 
	   \mu_{\psi, i}^{(k)}
	    \Big)^2}{
	    \sigma_{\eta, i}^{(k)}
	    }\bigg)\bigg],
	\end{split}
	\end{equation}
	where the $\lambda$ is a parameter that balances the influences of intra-space and inter-space objectives. 
	
	\subsection{GroupCL: GraphCL with Group Contrast}
	\label{sec: Group GraphCL}
    Built upon our framework, we extend the previous approach GraphCL into GroupCL, equipped with the proposed group contrastive objective. Figure~\ref{fig:framework} compares the frameworks of the original GraphCL and the extended variant. The two frameworks share the same data augmentation approach. Different from the graph-level encoder of GraphCL that only computes a single representation vector, the GroupCL encoder generates multiple vectors as representations. In particular, to control the computational cost, the generation of multiple representations shares the same node encoder $\mathcal{E}_n$ and involves a parameterized representor function $\mathcal{R}$ that computes multiple graph representations from node embeddings of a given graph.
    
    We first recap the computation of node embeddings in GraphCL, which is a common part shared by GraphCL and GroupCL. For both GraphCL and GroupCL, the computing procedures of views $u$ and $r$ are symmetric. We, therefore,  introduce the procedure of view $u$, and that of view $r$ is similar. For view $u$, GroupCL first performs a random data augmentation $\mathcal{T}_{u}$ on the input graph $(\bX, \bA)$ to generate the view $(\bX_u, \bA_u)$. The node encoder $\mathcal{E}_n$ then encodes each nodes of $(\bX_u, \bA_u)$ into node embeddings $\bU$. The mathematical formulations of these two steps are incorporated by
	\begin{equation}
	    \bU = \mathcal{E}_{n}(\mathcal{T}_{u}(\bX, \bA)).
	\end{equation}
	
\begin{figure}
		\includegraphics[width=\linewidth]{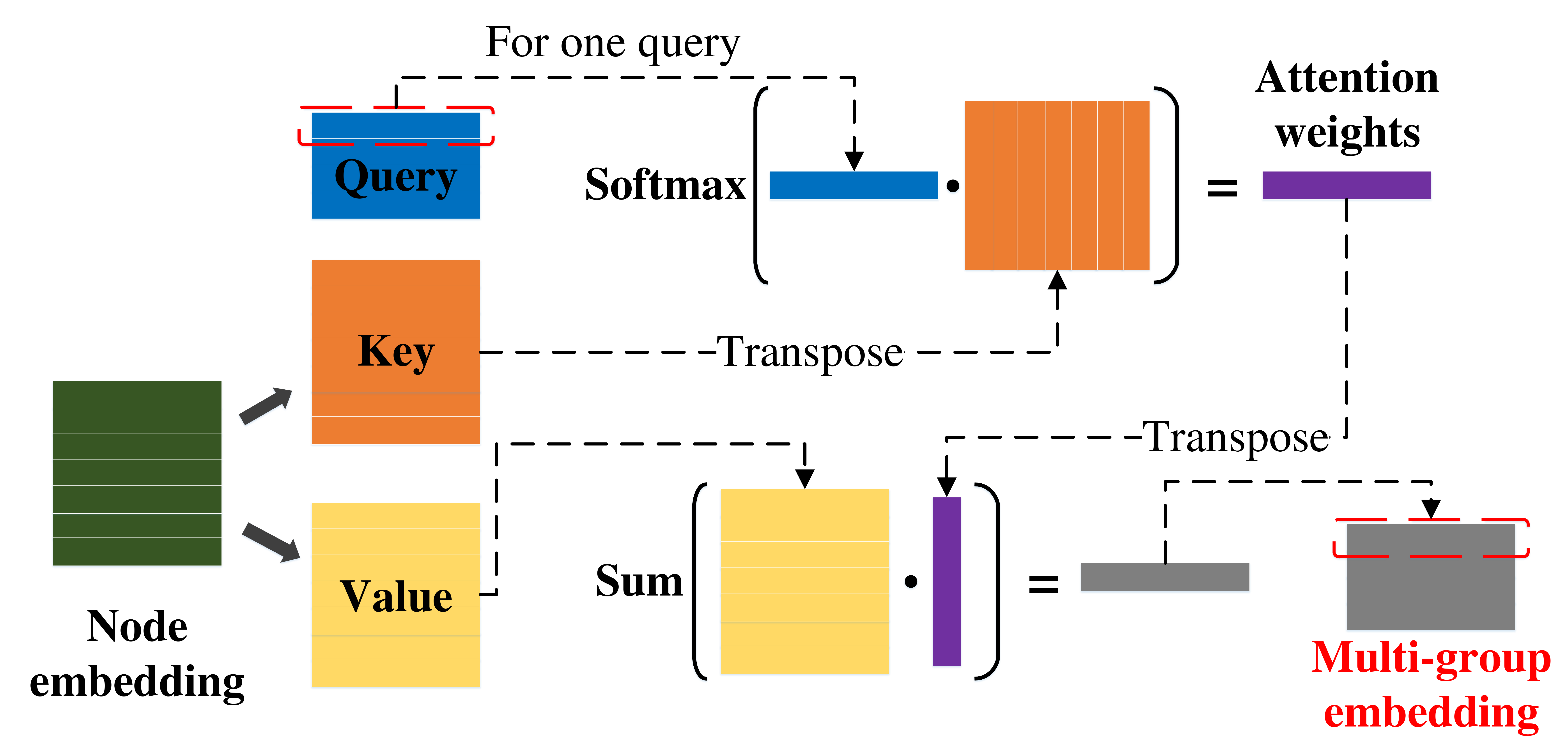}
		\caption{The computation pipeline of $\mathcal{R}_\mathbf{Q}$. It takes the node embedding matrix $\bU$ as input and outputs multiple graph-level embeddings. The node embedding matrix is firstly projected to the Key and Value matrices by two independent linear functions. Each vector in the Query matrix is involved in a dot product with the Key, leading to an attention weights vector. Then the attention vector is used to perform a weighted sum over the Value, generating one embedding.}
		\label{fig: Graph2Sub generator}
	\end{figure}

	Given node embeddings $\bU$, we take advantage of the attention mechanism to capture the information from different node combinations, where each graph representation depends greatly on the heavily attended nodes with respect to different queries. We hence instantiate the proposed representor function as $\mathcal{R}_\mathbf{Q}$ to
    capture information and encode diversified substructures of the graph. Figure~\ref{fig: Graph2Sub generator} demonstrates a complete computation pipeline of the proposed representor function $\mathcal{R}_\mathbf{Q}$.
	It employs two independent linear projections to map the node embedding matrix $\bU$ and outputs two matrices, the key matrix $\mathbf{K}$ and the value matrix $\mathbf{V}$. Mathematically,
	\begin{equation}
	\begin{split}
	\mathbf{K} = \bU \mathbf{W}_{K}   \in \mathbb{R}^{N \times d_{K}}, \\
	\mathbf{V} = \bU \mathbf{W}_{V}   \in \mathbb{R}^{N \times d_{V}},
	\end{split}	
	\end{equation}
	where $\mathbf{W}_{K} \in \mathbb{R}^{d_{n} \times d_{K}}$ and $\mathbf{W}_{V} \in \mathbb{R}^{d_{n} \times d_{V}}$ denote the projection matrices corresponding to the key and value, respectively, and $d_K$ and $d_V$ are the dimensions of them.
	
	The representor function $\mathcal{R}$ seeks to differently combine the nodes thereupon capture different graph substructures. To this end, we introduce a set of $p$ query vectors $\mathbf{Q} = [\mathbf{q}_{1}, \mathbf{q}_{2}, \cdots,  \mathbf{q}_{p}] \in \mathbb{R}^{d_{K} \times p}$, where each query induces one specific representation and the number of query vectors determines the number of groups in the proposed objective. In addition, the dimension of each query should be equal to the columns of $K$. This $\mathbf{Q}$ matrix is composed of trainable parameters which are randomly initialized and trained along with parameters in node encoders.
	The query $\mathbf{Q}$ is then employed to attend the key $\mathbf{K}$, producing the node-wise attention weights to be used for the computation of representations. A larger attention weight indicates a more informative node with respect to the corresponding query vector. We normalize the attention weights along the node dimension and finally perform the weighted summation over the value based on the attention weights to obtain the graph embedding. Mathematically, 
	\begin{equation}
	A = \text{Softmax}(\mathbf{K}\mathbf{Q}) \in \mathbb{R}^{N \times p},
	\end{equation}
	\begin{equation}
	[\mathbf{u}^{(1)}, \mathbf{u}^{(2)}, \cdots, \mathbf{u}^{(p)}] = \mathbf{V}^{T}\mathbf{A} \in \mathbb{R}^{d_{V} \times p},
	\label{eq:u}
	\end{equation}
	where we let $d_{V} = d_{o}/p$ as the $d_{V}$ determines the dimension of each embedding. 
	
	Given the representor function $\mathcal{R}_\mathbf{Q}$, the representations of the $u$ and $r$ views are computed as
	\begin{equation}
	\begin{split}
	    &[\mathbf{u}^{(1)}, \mathbf{u}^{(2)}, \cdots, \mathbf{u}^{(p)}] = \group\big(\mathcal{E}_{n}(\mathcal{T}_{u}(\bX, \bA)\big),\\
	    &[\mathbf{r}^{(1)}\ , \mathbf{r}^{(2)}\ , \cdots, \mathbf{r}^{(p)}] = \group\big(\mathcal{E}_{n}(\mathcal{T}_{r}(\bX, \bA)\big).
	\end{split}
	\end{equation}
	We then compute the group contrastive loss on representations $[\mathbf{u}^{(1)}, \mathbf{u}^{(2)}, \cdots, \mathbf{u}^{(p)}]$ and $[\mathbf{r}^{(1)}\ , \mathbf{r}^{(2)}\ , \cdots, \mathbf{r}^{(p)}]$ in groups by Equation~\eqref{eq: final_obj} and back propagate it to optimize the model.
    
    To sum up, $\mathcal{R}_\mathbf{Q}$ generates graph-level representations associated with different combinations of nodes, where each combination leads to individual substructural representation. Subject to our optimization objectives, the multiple representations are prompted to focus on different and informative combinations of nodes and thereupon encode different subgraph patterns.
    
	\renewcommand\arraystretch{1.4}
	\begin{table*}[!t]
		\centering
		\caption{The results of the unsupervised learning experiment. We run $10$-folder cross-validation and report the mean and the standard deviation of the classification accuracy. The best performance is highlighted by the bold number. The underlined numbers stand that the developed grouping methods are better than corresponding non-grouping methods.}
		\begin{tabular}{c|cccc|cccc}
			\hline
			Dataset&NCI1&PROTEINS&DD&MUTAG&COLLAB&RDT-B&RDT-M5K & IMDB-B \\
			\hline
			GL&-&-&-&81.66 $\pm$ 2.11&-&77.34 $\pm$ 0.18&41.01 $\pm$ 0.17&65.87 $\pm$ 0.98 \\
			WL&80.01 $\pm$ 0.50&72.92 $\pm$ 0.56&-&80.72 $\pm$ 3.00&-&68.82 $\pm$ 0.41&46.06 $\pm$ 0.21&72.30 $\pm$ 3.44 \\
			DGK&80.31 $\pm$ 0.46&73.30 $\pm$ 0.82&-&87.44 $\pm$ 2.72&-&78.04 $\pm$ 0.39&41.27 $\pm$ 0.18&66.96 $\pm$ 0.56 \\
			\hline
			node2vec&54.89 $\pm$ 1.61&57.49 $\pm$ 3.57&-&72.63 $\pm$ 10.20&-&-&-&- \\
			sub2vec&52.84  $\pm$ 1.61&53.03  $\pm$ 5.55&-&61.05 $\pm$ 15.80&-&71.48 $\pm$ 0.41&36.68 $\pm$ 0.42&55.26 $\pm$ 1.54 \\
			graph2vec&73.22 $\pm$ 1.81&73.30 $\pm$ 2.05&-&83.15 $\pm$ 9.25 &-&75.78 $\pm$ 1.03&47.86 $\pm$ 0.26&71.10 $\pm$ 0.54 \\
			\hline
			InfoGraph&76.20 $\pm$ 1.06&74.44 $\pm$ 0.31&72.85 $\pm$ 1.78&89.01 $\pm$ 1.13 & 70.65 $\pm$ 1.13&82.50 $\pm$ 1.42&53.46 $\pm$ 1.03&73.03 $\pm$ 0.87 \\
			GroupIG & 
			\underline{81.13 $\pm$ 0.73} & 
			\underline{74.74  $\pm$ 1.16} &
			\underline{73.67 $\pm$ 1.26} &
			\underline{89.82 $\pm$ 1.85} &
			\underline{76.18 $\pm$ 0.74} &
			\underline{90.55 $\pm$ 0.67} &
			\underline{54.72 $\pm$ 0.60} & 
			72.62 $\pm$ 0.78\\
			\hline
			GraphCL&77.87 $\pm$ 0.41&74.39 $\pm$ 0.45&\textbf{78.62 $\pm$ 0.40}&86.60 $\pm$ 1.34&71.36 $\pm$ 1.15&89.53 $\pm$ 0.84&\textbf{55.99 $\pm$ 0.28}&71.14 $\pm$ 0.44 \\
			GroupCL & 
			\underline{\textbf{81.69 $\pm$ 0.30}} & 
			\underline{\textbf{75.04 $\pm$ 0.54}} & 
			76.76 $\pm$ 1.20 & 
			\underline{\textbf{91.67 $\pm$ 1.37}} & 
			\underline{\textbf{76.38 $\pm$ 0.33}} & 
			\underline{\textbf{90.89 $\pm$ 0.85}} & 
			55.42 $\pm$ 0.22 & 
			\underline{\textbf{73.52 $\pm$ 0.95 }}\\
			\hline
		\end{tabular}
		\label{ul_results}
	\end{table*}
	
	\subsection{GroupIG: InfoGraph with Group Contrast}
	\label{sec: Group InfoGraph}
	In addition to the symmetric contrastive framework represented by GraphCL, our framework can be widely employed in any existing contrastive method for graph representation learning. To show the wide usability of our framework, we additionally employ it with InfoGraph~\cite{sun2019infograph} and propose the Group InfoGraph (GroupIG). In InfoGraph, representations of view $u$ and $r$ are the graph-level embedding and the node-level embedding, respectively. They are hence processed in an asymmetric manner. For view $u$, we use the same GNN encoder and representor function $\mathcal{R}_\mathbf{Q}$ as GroupCL. In particular, the multiple representations are obtained by
	\begin{equation}
	    [\mathbf{u}^{(1)}, \mathbf{u}^{(2)}, \cdots, \mathbf{u}^{(p)}] = \group\big(\mathcal{E}_{n}(\bX, \bA)\big).
	\end{equation}
	For view $r$, we generate the node embeddings through $\mathcal{E}_n$ and duplicate them for $p$ times to obtain $p$ representations, mathematically given by
	\begin{equation}
	    [\mathbf{r}^{(1)}, \mathbf{r}^{(2)}, \cdots, \mathbf{r}^{(p)}] = \text{Duplicate}(\mathcal{E}_{n}(\bX, \bA)).
	\end{equation}
	We compute the loss of $[\mathbf{u}^{(1)}, \mathbf{u}^{(2)}, \cdots, \mathbf{u}^{(p)}]$ and $[\mathbf{r}^{(1)}\ , \mathbf{r}^{(2)}\ , \cdots, \mathbf{r}^{(p)}]$ by Equation~\eqref{eq: final_obj} and back propagate it to optimize the model.

	\section{Experimental Studies}
	\label{sec: ES}
	In this section, we assess the effectiveness of our group contrastive learning framework on both graph unsupervised classification and graph transfer learning tasks. To illustrate the property of groups, we also investigate the correlation between different groups and visualize the content captured by different groups in Section~4.3. Furthermore, we conduct comprehensive ablation studies to demonstrate the influence of various hyperparameters, including the number of groups $p$, the weight of diversity loss $\lambda$, and two different types of MI upper bound estimators in Section 4.4. Finally, we analyze the training complexity of the grouping method and non-grouping methods in Section~4.5.

	Our implementation is based on Pytorch~\cite{NEURIPS2019_9015} and Pytorch Geometric~\cite{Fey/Lenssen/2019} libraries. The Adam optimizer is adopted for the optimization of the model. We set the number of groups to $4$, which means that we generate $4$ feature vectors for each graph example. The total embedding dimension is set to $160$ for these two tasks, and the dimension for each representation is determined by the total embedding dimension dividing the number of groups. We adopt Equation~\eqref{eq: final_obj} as the objective function for our main results in Section~4.1 and Section~4.2, since the correspondence exists between the same dimension across multiple representations generated by $\group(\cdot)$. 
	The objective function introduces one additional hyper-parameter $\lambda$, of which the value is chosen from $[0.1, 0.3, 0.5, 0.7]$.  
	
	\renewcommand\arraystretch{1.2}
	\begin{table*}[!t]
		\centering
		\caption{The results of the transfer learning experiment. We run $10$-folder cross-validation and report the mean and the standard deviation of the ROC-AUC scores. The best performance is highlighted by the bold number. The underlined numbers stand that the developed grouping methods are better than corresponding non-grouping methods.}
		\begin{tabular}{c|cccccccc}
			\hline
			Dataset&BBBP&Tox21&ToxCast&SIDER&ClinTox&MUV&HIV&BACE  \\
			\hline
			No Pre-Train&65.8 $\pm$ 4.5&74.0 $\pm$ 0.8&63.4 $\pm$ 0.6&57.3 $\pm$ 1.6&58.0 $\pm$ 4.4&71.8 $\pm$ 2.5&75.3 $\pm$ 1.9&70.1 $\pm$ 5.4 \\
			Infomax&68.8 $\pm$ 0.8&75.3 $\pm$ 0.5&62.7 $\pm$ 0.4&58.4 $\pm$ 0.8&69.9 $\pm$ 3.0&75.3 $\pm$ 2.5&76.0 $\pm$ 0.7&75.9 $\pm$ 1.6 \\
			EdgePred&67.3 $\pm$ 2.4&76.0 $\pm$ 0.6&64.1 $\pm$ 0.6&60.4 $\pm$ 0.7&64.1 $\pm$ 3.7&74.1 $\pm$ 2.1&76.3 $\pm$ 1.0&79.9 $\pm$ 0.9 \\
			AttrMasking&64.3 $\pm$ 2.8&\textbf{76.7 $\pm$ 0.4}&\textbf{64.2 $\pm$ 0.5}&61.0 $\pm$ 0.7&71.8 $\pm$ 4.1&74.7 $\pm$ 1.4&77.2 $\pm$ 1.1&79.3 $\pm$ 1.6 \\
			ContextPred&68.0 $\pm$ 2.0 &75.7 $\pm$ 0.7&63.9 $\pm$ 0.6&60.9 $\pm$ 0.6&65.9 $\pm$ 3.8&\textbf{75.8 $\pm$ 1.7}&77.3 $\pm$ 1.0&79.6 $\pm$ 1.2\\
			\hline
			GraphCL&69.7 $\pm$ 0.7&73.9 $\pm$ 0.7&62.4 $\pm$ 0.6&60.5 $\pm$ 0.9&76.0 $\pm$ 2.7&69.8 $\pm$ 2.7&\textbf{78.5 $\pm$ 1.2}&75.4 $\pm$ 1.4 \\
			GroupCL & 
			\underline{\textbf{71.04 $\pm$ 1.25}}&
			\underline{75.47 $\pm$ 0.40}&
			\underline{62.66 $\pm$ 0.95}&
			\underline{\textbf{61.48 $\pm$ 0.90}}&
			\underline{\textbf{80.90 $\pm$ 2.86}}&
			\underline{73.22 $\pm$ 2.25}&
			76.68 $\pm$ 1.17&
			\underline{\textbf{80.95 $\pm$ 1.88}} \\
			\hline
		\end{tabular}
		\label{tl_results}
	\end{table*}
	\subsection{Unsupervised Learning}
	We evaluate our proposed framework with unsupervised graph classification tasks. Following the learning procedure in~\cite{you2020graph}, we first train the GNN model in a self-supervised fashion. The trained model is then fixed and used to generate the graph embeddings during the evaluation phase. Finally, an SVM is adopted to classify the embeddings into different categories. 
	
	\textbf{Datasets and baselines.} In this experiment, two types of datasets are considered, those are the biochemical molecules and the social networks. The statistics information is summarized in Table~\ref{ul_datasets}. 
	We compare our GroupCL and GroupIG with the original approaches as baselines, \textit{i.e.}, GraphCL and InfoGraph, respectively. 
	Additionally, we compare our grouping methods with other six SOTA graph unsupervised learning methods, including graphlet kernel (GL), Weisfeiler-Lehman sub-tree kernel (WL), deep graph kernel (DGK), node2vec~\cite{grover2016node2vec}, sub2vec~\cite{adhikari2018sub2vec}, and graph2vec\cite{narayanan2017graph2vec}.  
	
	\textbf{Experimental configurations.} For the node encoder $\mathcal{E}_{n}$, we build a three-layer the graph isomorphism network (GIN)~\cite{xu2018powerful} with $32$ hidden units. 
	To a fair comparison, the total embedding dimensions are to the same number $160$ for both grouping methods and their baselines. The Adam algorithm with a learning rate of 0.001 to optimizing the GNN model. We follow the GraphCL~\cite{you2020graph} to choose the number of epochs as $20$, batch size as $128$, and the C parameter of SVM from $[10^{-3}, 10^{-2}, ..., 10^2, 10^3]$. 
	We closely follow the evaluation settings adopted by the previous state-of-the-art approaches~\cite{you2020graph, hassani2020contrastive}. In particular, we split the dataset into the train, test, and validation sets at the proportion of $8:1:1$ and report the mean classification accuracy with standard deviation after 5 runs followed by a linear SVM classifier. The SVM is trained using cross-validation on training folds of data and the model for testing is selected by the best validation performance.
	
	\textbf{Results.} Table~\ref{ul_results} shows the results of the unsupervised learning experiments. Shown numbers of the baseline methods come from GraphCL~\cite{you2020graph}.
	Numbers are highlighted with underlines when grouping methods obtained better performance than the original ones. It shows that GroupIG outperforms InfoGraph on seven out of eight datasets, and GroupCL achieves better performance on six out of eight datasets than GraphCL. In some datasets, our grouping methods outperform the original version with a large margin. For example, GroupIG obtained about $5 \%$, $6 \%$, and $8 \%$ higher accuracy than InfoGraph, on the NCI1, COLLAB, and RDT-B respectively. GroupCL achieved about $4 \%$, $5 \%$, and $5 \%$ higher accuracy than GraphCL, on the NCI1, MUTAG, and COLLAB correspondingly. The results substantially demonstrate the effectiveness of the grouping technique. We then compare the results of our methods with the SOTA performance achieved by the previous methods. The best results among all methods are in bold. GroupCL achieves new SOTA accuracy on six out of eight datasets and the second-best accuracy on the rest two datasets.
	Moreover, our method outperforms the state-of-the-art approaches at significant margins on some datasets, like about $5\%$ on COLLAB, $3\%$ on MUTAG. To sum up, our grouping methods outperform the previous single-embedding one consistently and significantly. We will further verify it by ablation study on the number of groups in Section~\ref{sec: AS}.
	
	\subsection{Transfer Learning}
	Furthermore, we conduct an evaluation with transfer learning on molecular chemical property prediction tasks. Specifically, we pre-train the model on a large-scale unlabeled dataset, fine-tune and evaluate the model on multiple different downstream datasets. The goal is to evaluate the transferability of models trained under different pre-training schemes.
	
	\textbf{Datasets and baselines.} We adopt ZINC-2M for pre-training, which contains 2 million unlabeled molecules sampled from the ZINC15 database~\cite{sterling2015zinc}. For downstream tasks, we include $8$ binary classification datasets provided by MoleculeNet~\cite{wu2018moleculenet}. The concrete information is given in Table~\ref{tl_datasets}. For baseline methods, we include results of various pre-training strategies, including Infomax, EdgePred, AttrMasking, and ContextPred, provided in~\cite{hu2019strategies} as well as results without pre-training.
	
	\textbf{Experimental configurations.} We adopt the same GNN encoder in this experiment to the one used in the unsupervised learning experiments. Following~\cite{hu2019strategies}, the batch size and training epochs are set to $256$ and $100$ for pre-training, then $32$ and $50$ for finetuning. For both pretraining and finetuning, the learning rate is fixed to $0.001$.
	The same data splitting method used in unsupervised learning is applied to the transfer learning evaluation. We conduct $10$ times cross-validation experiments and report the mean and standard deviation of ROC-AUC scores in Table~\ref{tl_results}.
	
	\textbf{Results.}
	Table~\ref{tl_results} shows the comparison results with baselines. Our method achieves state-of-the-art performance on BBBP, SIDER, ClinTox, and BACE. It is worth noting that our method is $4.9\%$ higher than the previous best performance on ClinTox. In this task, the best performance of each dataset distributes on the baseline methods in a decentralized manner as different downstream tasks differ significantly. However, our method, in general, achieves the best performance on more datasets than the previous methods do: the state-of-the-art performances of four datasets are achieved by our methods, that of two datasets achieved by AttrNasking, and that of the other two datasets achieved by the ContextPred and GraphCL respectively. Regarding the comparison with our direct baseline GraphCL, our method obtains better results on 7 in 8 datasets. This comparison indicates the effectiveness of our proposed framework.
	
    \begin{figure}[t!]
	    \centering
		\includegraphics[width=0.98\linewidth]{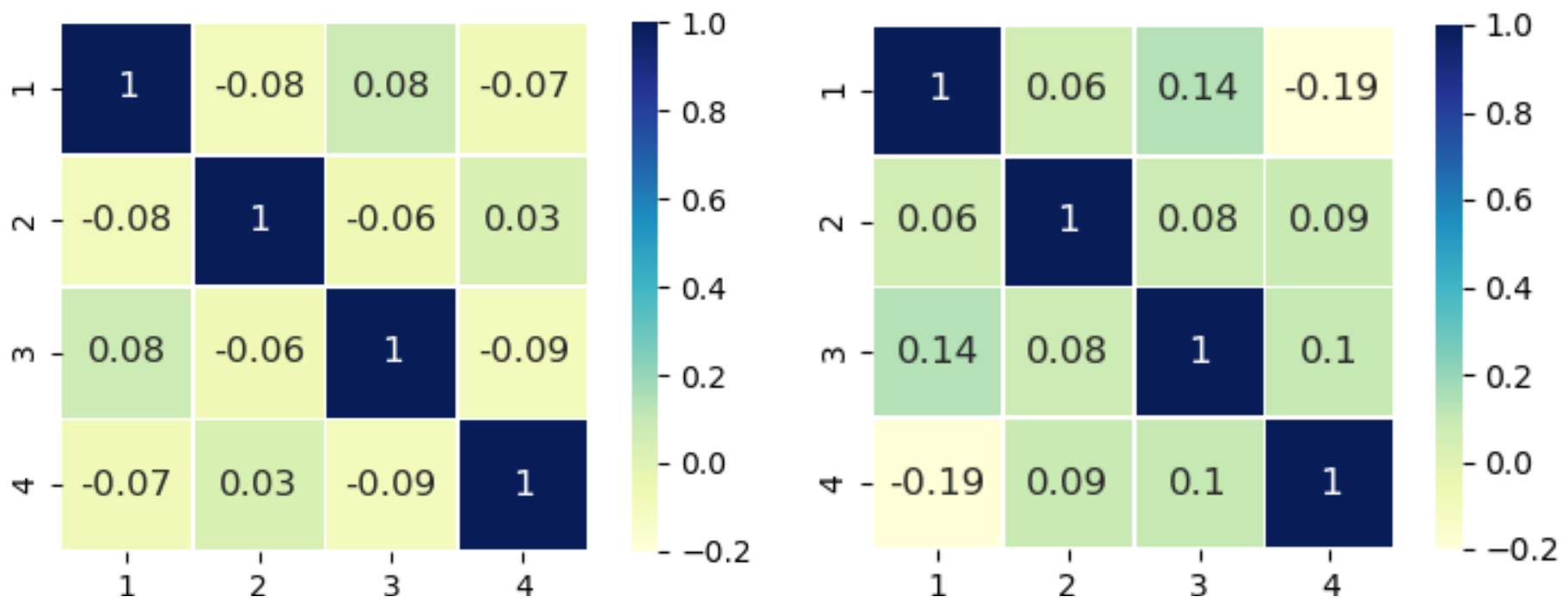}
		\caption{The cosine distance of queries of PROTEINS dataset (left) and MUTAG dataset (right) on the unsupervised learning task.}
		\label{fig: group_cor}
	\end{figure}

	\subsection{Study of Groups}
	Our group contrastive learning framework performs contrastive learning under multiple measures, making each representation capture one specific characteristic. To support this claim, we conduct the quantitative representation correlation study and the qualitative representation attention study. 

	\begin{figure*}[hbt!]
	    \centering
		\includegraphics[width=0.9\linewidth]{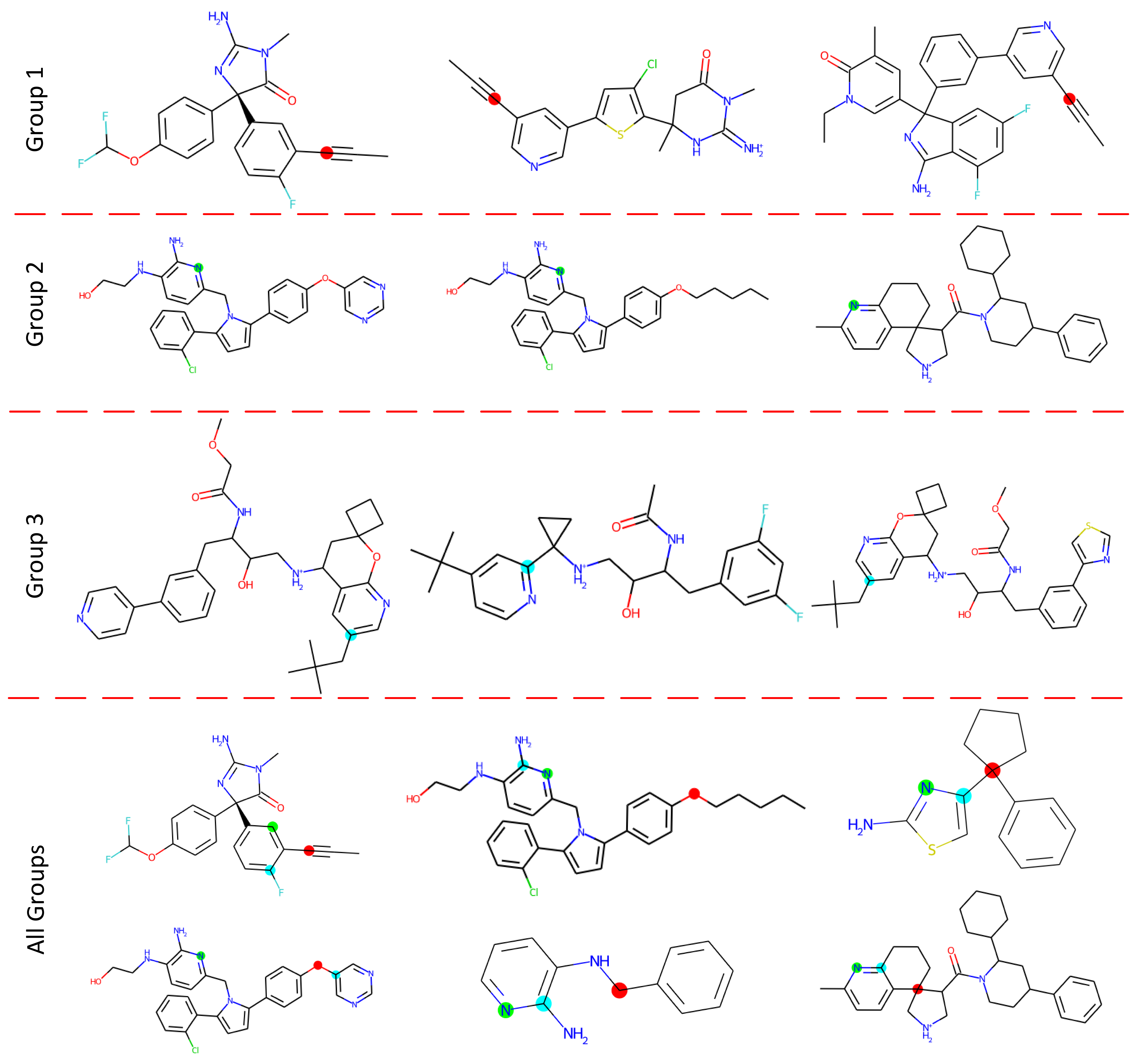}
		\caption{Visualization of attention weights of BACE dataset on the transfer learning task. The first three rows exhibit the independent results of three representations, in which the nodes being paid large attention are highlighted by the red, green, and cyan circles correspondingly. The last row shows the results of all-representation attentions on one graph.}
		\label{fig: vis_results}
	\end{figure*}
	\renewcommand\arraystretch{1.2}
	\begin{table*}[!t]
		\centering
		\caption{Ablation study on MI upper bound estimator. We run $10$-folder cross-validation and report the mean and the standard deviation of the classification accuracy. The better performance is highlighted by the bold number.}
		\begin{tabular}{c|c|c|c|c}
			\hline
			$\text{CLUB Estimators}$ & NCI1 & MUTAG & COLLAB & IMDB-B \\
			\hline
			No CLUB 
			& 77.87 $\pm$ 0.41
			& 86.60 $\pm$ 1.34 
			& 71.36 $\pm$ 1.15 
			& 71.14 $\pm$ 0.44 \\
			\hline
			 Parameterized
			 & 79.87 $\pm$ 0.59
			 & 90.63 $\pm$ 1.07
			 & 75.18 $\pm$ 0.88
			 & 73.38 $\pm$ 0.73 \\
			 Non-parameterized 
			 & \textbf{81.69 $\pm$ 0.30}
			 & \textbf{91.67 $\pm$ 1.37}
			 & \textbf{76.38  $\pm$ 0.33} 
			 & \textbf{73.52 $\pm$ 0.95} \\
			\hline
		\end{tabular}
		\label{exp: as_mie}
	\end{table*}
	
	\textbf{Group correlation.} We investigate the correlation of representations across different groups to demonstrate that diversities among them. We conduct this study on the PROTEINS dataset and the MUTAG dataset in the unsupervised learning experiment. We set the group to $4$.
	As introduced in Section~\ref{sec: Group GraphCL}, the groups of representations are generated by employing attention with different query vectors. Hence, we compute the cosine distance between two queries as the correlation strength between each pair of groups. Figure~\ref{fig: group_cor} shows the results in a matrix, in which diagonal elements are the cosine distances of the same queries, and the other elements are those of different queries. We can observe that the cosine distance between any two different queries is quite small, implying that the correlation between any two different groups is weak.

	\textbf{Attention maps of different groups.} We study the attention weights to reveal what kind of characteristics do representations across different groups encode. Specifically, we find the nodes which have the largest attention weights for each representation and visualize the graph that those nodes belong to. We conduct this study on the BACE dataset in the transfer learning experiment, in which the number of groups is set to $3$. Figure~\ref{fig: vis_results} demonstrate the attentions of three representations independently, as well as those of three representations on one graph. Nodes that have larger attention weights are highlighted by the red, green, and cyan circles respectively for those three groups. The three rows on top show that one representation consistently focuses on the same nodes or motifs among different molecule instances and different representations capture information of different substructures. The above results support our claim that representations learned with our framework can encode diverse characteristics of given graphs.
	Putting visualization results of all representations together and studying more examples, the bottom row in Figure~\ref{fig: vis_results} indicates that one representation is not trivially limited to a single type of node or substructure. In particular, when the corresponding type of node does not appear in a graph, a representation is also able to capture nodes in similar substructures.
	
	\subsection{MI Upper Bound Estimators}
	We study the performance of two MI upper bound estimators introduced in Section~\ref{sec: inter-group}, which are the non-parameterized CLUB and the parameterized CLUB. We conduct this experiment on two small datasets: MUTAG and IMDB-B, and two large datasets: NCI1 and COLLAB in the unsupervised learning task. Table~\ref{exp: as_mie} shows the results, in which we observe that the non-parameterized CLUB outperforms the parameterized one on the four datasets with our developed representor function. Two potential causes lead to these results. First, the correspondence exists among the same dimensions across different representations generated by our representor function. The non-parameterized CLUB is suitable for this case and can well estimate the lower bound of MI. 
	Furthermore, compared to the non-parameterized estimator, the parameterized CLUB introduces additional complexity to the model which makes the learning process more difficult. 
	
	However, models optimized with parameterized CLUB still consistently outperforms the baseline methods without group contrast. In cases when the correspondence does not exist among the same dimensions across different groups and the non-parameterized CLUB becomes inapplicable, one can still adopt the parameterized estimator that significantly improves the baseline performance.
	
	\renewcommand\arraystretch{1.2}
	\begin{table}[!t]
		\centering
		\caption{Ablation study on the number of groups $p$. We run $10$-folder cross-validation and report the mean and the standard deviation of the classification accuracy. The best performance is highlighted by the bold number.}
		\begin{tabular}{c|c|c|c|c}
			\hline
			$p$ & PROTEINS & MUTAG & DD & IMDB-B    \\
			\hline
			 1& 73.64 $\pm$ 0.81 & 90.85 $\pm$ 1.84  &73.92 $\pm$ 0.42 & 71.28 $\pm$ 0.74\\
			 2&73.98 $\pm$ 0.50&90.20 $\pm$ 1.69&74.98 $\pm$ 0.99&71.90 $\pm$ 1.64 \\
			 3&74.25 $\pm$ 0.39&89.68 $\pm$ 0.99&74.84 $\pm$ 0.87& 71.98 $\pm$ 0.88\\
			 4&74.67 $\pm$ 0.74 &\textbf{91.67 $\pm$ 1.37}& \textbf{76.76 $\pm$ 1.20}&\textbf{73.52 $\pm$ 0.95}\\
			 5&\textbf{75.04 $\pm$ 0.54} & 90.08 $\pm$ 2.52 &75.87 $\pm$ 0.45 &70.64 $\pm$ 2.11\\
			\hline
		\end{tabular}
		\label{exp: as_p}
	\end{table}
	
	\renewcommand\arraystretch{1.2}
	\begin{table}[!t]
		\centering
		\caption{Ablation study on diversity loss weight $\lambda$. We run $10$-folder cross-validation and report the mean and the standard deviation of the classification accuracy. The best performance is highlighted by the bold number.}
	\begin{tabular}{c|c|c|c|c}
			\hline
			$\lambda$ & PROTEINS & MUTAG & DD & IMDB-B \\
			\hline
			 0.0&73.03 $\pm$ 1.05&90.52 $\pm$ 1.88&74.82 $\pm$ 0.65& 71.96 $\pm$ 1.23 \\
			 0.1&74.64 $\pm$ 1.21&91.37 $\pm$ 1.44&75.19  $\pm$ 0.37& 72.32 $\pm$ 1.35\\
			 0.3&74.16 $\pm$ 0.97&91.37 $\pm$ 1.03&75.43 $\pm$ 0.40 & 72.53 $\pm$ 0.99\\
			 0.5& 74.59 $\pm$ 0.47& \textbf{91.67 $\pm$ 1.37} &\textbf{76.76 $\pm$ 1.20}& \textbf{73.52 $\pm$ 0.95} \\
			 0.7&\textbf{75.04$\pm$ 0.54}&91.47 $\pm$1.12 &75.63 $\pm$ 0.44& 72.42  $\pm$ 1.23\\
			 0.9&74.32 $\pm$ 0.70&91.37 $\pm$ 1.02&75.82 $\pm$ 0.27& 71.95 $\pm$ 1.20\\
			\hline
		\end{tabular}
		\label{exp: as_lambda}
	\end{table}
	
	\subsection{Ablation Studies}
	\label{sec: AS}
	We conduct ablation studies to explore the effect of critical hyperparameters, including the number of groups and the weight $\lambda$ that balances the influence between the intra-space objective and inter-space objective. We consider four smaller datasets in all the eight ones, which pursues high efficiency. The ablation studies are conducted in the unsupervised learning setting. 
	
	\textbf{Number of groups: $p$.} Compared to the prior work, the main breakthrough made by our approach is to learn multiple graph-level embeddings. Therefore, the number of groups: $p$ is a vital parameter to demonstrate the effectiveness of our approach. Through the studies of $p$, we want to find the difference between the results of single group and multi-group, as well as the proper number of groups in the multi-group case. To this end, we set $p=\{1, 2, 3, 4, 5\}$.
	Table~\ref{exp: as_p} gives the results. We achieve the best performance on PROTEINS when $p=5$ and on the other three datasets when $p=4$. 
	Note that in most cases the performance of $p=1$ is worse than that of $p>1$, which demonstrates the effectiveness of the grouping scheme.
	
	\textbf{Weight of diversity loss: $\lambda$.} There are two components in objectives employed to optimize our model, the intra-space terms and the inter-space terms. To balance their effects, we employ a hyperparameter $\lambda$ to weight the inter-space objective. We compare the values varying from $\{0.0, 0.1, 0.3, 0.5, 0.7, 0.9\}$. Note that $\lambda=0.0$ indicates that no diversity is enforced among groups. We hence let $\lambda=0.0$ serve as the controlled experiment, which is used to justify the necessity of the diversities of groups. The results are shown in Table~\ref{exp: as_lambda}. We observe that the results are consistently worse when $\lambda=0.0$, which demonstrates that the diversities of groups are crucial for boosting performance. For those four datasets, the best performance is obtained when $\lambda$ equals $0.7$, $0.5$, $0.5$, and $0.5$ accordingly.

	\renewcommand\arraystretch{1.2}
	\begin{table}[!t]
		\centering
		\caption{Comparison of the number of parameters. The number of parameters excludes the parameters in GNN which is shared by GraphCL and GroupCL.}
		\begin{tabular}{c|c|c}
			\hline
			Methods & GroupCL & GraphCL \\
			\hline
			No. of Params & 22,800 & 51,200 \\
			\hline
		\end{tabular}
		\label{exp: complexity}
	\end{table}	
	
    \subsection{Complexity Study}
    Taking GroupCL as an example, we analyze the training complexity of our grouping framework. We compare it with the non-grouping method GraphCL. According to Figure~\ref{fig:framework}, GroupCL and GraphCL share the same GNN, so we only need to explore the complexity of the operations after the GNN, \textit{i.e.}, the representor function and projection heads. GraphCL sets the dimension of the node embedding to $160$ and that of the graph embedding to $160$ as well. We follow this setting. For the hyperparameters involved in GroupCL, we set the number of groups to $4$ and the dimension of the Key matrix to $100$. Thus we have that the Query matrix is of size $4 \times 100$, $\mathbf{W}_{K}$ is of size $160 \times 100$, and $\mathbf{W}_{V}$ is of size $160 \times (160 / 4)$. The projection head of GraphCL is a neural network of two fully connected layers, each of which has $160$ units. So the number of parameters is $2 \times 160 \times 160$. Table~\ref{exp: complexity} summarizes the results, and we can observe that GroupCL has fewer parameters than GraphCL.

	\renewcommand\arraystretch{1.2}
	\begin{table*}[ht]
		\centering
		\caption{The statistics of the datasets used in the unsupervised learning experiment.}
		\begin{tabular}{c|c|c|c|c|c|c}
			\hline
			Datasets& Type &\#Graphs&\#Graph Classes&\#Avg. Nodes&\#Avg. Edges& \#Node Classes  \\
			\hline
			NCI1 &Chemistry&4110&2&29.9&32.3&37 \\
			Proteins &Chemistry&1113&2&39.1&72.8&3 \\
			Mutag &Chemistry&188&2&17.9&19.8&7 \\
			DD &Chemistry&1178&2&284.3&725.7&89 \\
			\hline
			COLLAB &Social&5000&3&74.5&2457.8&- \\
			IMDB-B &Social&1000&2&19.8&96.5&-\\
			Reddit-B &Social&2000&2&429.6&497.8&- \\
			Reddit-5K &Social&4999&5&508.8&594.9&- \\
			\hline
		\end{tabular}
		\label{ul_datasets}
	\end{table*}
	
	\renewcommand\arraystretch{1.2}
	\begin{table*}[ht]
		\centering
		\caption{The statistics of the datasets used in the transfer learning experiment.}
		\begin{tabular}{c|c|c|c|c|c|c}
			\hline
			Datasets& Type &\#Graphs&\#Graph Classes&\#Avg. Nodes&\#Avg. Edges& \#Binary prediction tasks  \\
			\hline
			ZINC-2M&Molecule, Pre-training&2,000,000&-&25.5&27.5&1\\
			\hline
			BBBP&Molecule, Finetuning&2039&2&24.1&51.9&1\\
			Tox21&Molecule, Finetuning&7831&2&18.6&38.6&12\\
			ToxCast&Molecule, Finetuning&8575&2&18.8&18.8&617\\
			SIDER&Molecule, Finetuning&1427&2&33.6&70.7&27\\
			Clintox&Molecule, Finetuning&1478&2&26.1&55.7&2\\
			MUV&Molecule, Finetuning&93087&2&24.2&52.6&17\\
			HIV&Molecule, Finetuning&41127&2&25.5&54.9&1\\
			BACE&Molecule, Finetuning&1513&2&34.1&73.7&1\\
			\hline
		\end{tabular}
		\label{tl_datasets}
	\end{table*}
	
\section{Conclusions and Outlook}
In this work, we have proposed a group contrastive learning framework to perform unsupervised graph representation learning. 
Our framework is more powerful than most previous contrastive learning methods, since it contrasts multiple representations in various subspaces, thereby enables these representations to encode abundant characteristics of graphs. 
To learn informative and diverse representations, we have developed two principled objectives regarding the intra-space and the inter-space. 
We have further proposed an attention-based representor function to incorporate into our framework. The generated representations by this function are capable of encoding informative and diverse graph substructures.  
Built on our framework, we have extended two prior non-grouping methods into GroupCL and GroupIG. To verify their effectiveness, we have conducted thorough experiments on graph unsupervised learning and transfer learning tasks. The quantitative results demonstrate that our methods achieve new state-of-the-art performance on a majority datasets and outperform the non-grouping methods obviously and consistently. In particular, we have visualized the attention weights of each node, which are learned by our representor function. The qualitative results illustrate that these representations focus on important and diverse graph substructures. 

Currently, we consider one type of function, inserted after the GNN, to generate multiple representations.  In the future, we plan to explore more multi-representation generating manners and consider different inserting positions.

	\bibliography{GroupCL}

\begin{thebibliography}{10}
\providecommand{\url}[1]{#1}
\csname url@samestyle\endcsname
\providecommand{\newblock}{\relax}
\providecommand{\bibinfo}[2]{#2}
\providecommand{\BIBentrySTDinterwordspacing}{\spaceskip=0pt\relax}
\providecommand{\BIBentryALTinterwordstretchfactor}{4}
\providecommand{\BIBentryALTinterwordspacing}{\spaceskip=\fontdimen2\font plus
\BIBentryALTinterwordstretchfactor\fontdimen3\font minus
  \fontdimen4\font\relax}
\providecommand{\BIBforeignlanguage}[2]{{%
\expandafter\ifx\csname l@#1\endcsname\relax
\typeout{** WARNING: IEEEtran.bst: No hyphenation pattern has been}%
\typeout{** loaded for the language `#1'. Using the pattern for}%
\typeout{** the default language instead.}%
\else
\language=\csname l@#1\endcsname
\fi
#2}}
\providecommand{\BIBdecl}{\relax}
\BIBdecl

\bibitem{kipf2016semi}
T.~N. Kipf and M.~Welling, ``Semi-supervised classification with graph
  convolutional networks,'' in \emph{International Conference on Learning
  Representations}, 2017.

\bibitem{gao2019graph}
H.~Gao and S.~Ji, ``Graph representation learning via hard and channel-wise
  attention networks,'' in \emph{Proceedings of the 25th ACM SIGKDD
  International Conference on Knowledge Discovery and Data Mining}, 2019, pp.
  741--749.

\bibitem{velivckovic2017graph}
P.~Veli{\v{c}}kovi{\'c}, G.~Cucurull, A.~Casanova, A.~Romero, P.~Lio, and
  Y.~Bengio, ``Graph attention networks,'' in \emph{International Conference on
  Learning Representations}, 2018.

\bibitem{liu2020deep}
Y.~Liu, H.~Yuan, L.~Cai, and S.~Ji, ``Deep learning of high-order interactions
  for protein interface prediction,'' in \emph{Proceedings of the 26th ACM
  SIGKDD International Conference on Knowledge Discovery and Data Mining},
  2020, pp. 679--687.

\bibitem{gao2018large}
H.~Gao, Z.~Wang, and S.~Ji, ``Large-scale learnable graph convolutional
  networks,'' in \emph{Proceedings of the 24th ACM SIGKDD International
  Conference on Knowledge Discovery and Data Mining}, 2018, pp. 1416--1424.

\bibitem{yuan2020structpool}
H.~Yuan and S.~Ji, ``Structpool: Structured graph pooling via conditional
  random fields,'' in \emph{International Conference on Learning
  Representations}, 2020.

\bibitem{rong2020self}
Y.~Rong, Y.~Bian, T.~Xu, W.~Xie, Y.~Wei, W.~Huang, and J.~Huang,
  ``Self-supervised graph transformer on large-scale molecular data,'' in
  \emph{Advances in Neural Information Processing Systems}, 2020.

\bibitem{wang2020second}
Z.~Wang and S.~Ji, ``Second-order pooling for graph neural networks,''
  \emph{IEEE Transactions on Pattern Analysis and Machine Intelligence}, 2020.

\bibitem{xie2020noise2same}
Y.~Xie, Z.~Wang, and S.~Ji, ``Noise2same: Optimizing a self-supervised bound
  for image denoising,'' in \emph{Advances in Neural Information Processing
  Systems}, 2020.

\bibitem{wang2021contrastive}
X.~Wang and G.-J. Qi, ``Contrastive learning with stronger augmentations,''
  \emph{arXiv preprint arXiv:2104.07713}, 2021.

\bibitem{you2020graph}
Y.~You, T.~Chen, Y.~Sui, T.~Chen, Z.~Wang, and Y.~Shen, ``Graph contrastive
  learning with augmentations,'' in \emph{Advances in Neural Information
  Processing Systems}, vol.~33, 2020.

\bibitem{chen2020simple}
T.~Chen, S.~Kornblith, M.~Norouzi, and G.~Hinton, ``A simple framework for
  contrastive learning of visual representations,'' in \emph{International
  Conference on Machine Learning}, 2020, pp. 1597--1607.

\bibitem{pathak2016context}
D.~Pathak, P.~Krahenbuhl, J.~Donahue, T.~Darrell, and A.~A. Efros, ``Context
  encoders: Feature learning by inpainting,'' in \emph{Proceedings of the IEEE
  Conference on Computer Vision and Pattern Recognition}, 2016, pp. 2536--2544.

\bibitem{doersch2015unsupervised}
C.~Doersch, A.~Gupta, and A.~A. Efros, ``Unsupervised visual representation
  learning by context prediction,'' in \emph{Proceedings of the IEEE
  International Conference on Computer Vision}, 2015, pp. 1422--1430.

\bibitem{devlin2018bert}
J.~Devlin, M.-W. Chang, K.~Lee, and K.~Toutanova, ``Bert: Pre-training of deep
  bidirectional transformers for language understanding,'' in \emph{North
  American Chapter of the Association for Computational Linguistics}, 2019.

\bibitem{wang2019self}
H.~Wang, X.~Wang, W.~Xiong, M.~Yu, X.~Guo, S.~Chang, and W.~Y. Wang,
  ``Self-supervised learning for contextualized extractive summarization,'' in
  \emph{Proceedings of the 57th Annual Meeting of the Association for
  Computational Linguistics}, 2019.

\bibitem{yang2019xlnet}
Z.~Yang, Z.~Dai, Y.~Yang, J.~Carbonell, R.~R. Salakhutdinov, and Q.~V. Le,
  ``Xlnet: Generalized autoregressive pretraining for language understanding,''
  in \emph{Advances in Neural Information Processing Systems}, vol.~32, 2019.

\bibitem{sun2019infograph}
F.-Y. Sun, J.~Hoffmann, V.~Verma, and J.~Tang, ``Infograph: Unsupervised and
  semi-supervised graph-level representation learning via mutual information
  maximization,'' in \emph{International Conference on Learning
  Representations}, 2020.

\bibitem{thakoor2021bootstrapped}
S.~Thakoor, C.~Tallec, M.~G. Azar, R.~Munos, P.~Veli{\v{c}}kovi{\'c}, and
  M.~Valko, ``Bootstrapped representation learning on graphs,'' \emph{arXiv
  preprint arXiv:2102.06514}, 2021.

\bibitem{zhu2020deep}
Y.~Zhu, Y.~Xu, F.~Yu, Q.~Liu, S.~Wu, and L.~Wang, ``Deep graph contrastive
  representation learning,'' in \emph{International Conference on Machine
  Learning Workshops}, 2020.

\bibitem{jiao2020sub}
Y.~Jiao, Y.~Xiong, J.~Zhang, Y.~Zhang, T.~Zhang, and Y.~Zhu, ``Sub-graph
  contrast for scalable self-supervised graph representation learning,'' in
  \emph{IEEE International Conference on Data Mining}, 2020, pp. 222--231.

\bibitem{peng2020graph}
Z.~Peng, W.~Huang, M.~Luo, Q.~Zheng, Y.~Rong, T.~Xu, and J.~Huang, ``Graph
  representation learning via graphical mutual information maximization,'' in
  \emph{Proceedings of the Web Conference}, 2020, pp. 259--270.

\bibitem{velivckovic2018deep}
P.~Veli{\v{c}}kovi{\'c}, W.~Fedus, W.~L. Hamilton, P.~Li{\`o}, Y.~Bengio, and
  R.~D. Hjelm, ``Deep graph infomax,'' in \emph{International Conference on
  Learning Representations}, 2019.

\bibitem{hamilton2020graph}
W.~L. Hamilton, ``Graph representation learning,'' \emph{Synthesis Lectures on
  Artifical Intelligence and Machine Learning}, vol.~14, no.~3, pp. 1--159,
  2020.

\bibitem{wang2017mgae}
C.~Wang, S.~Pan, G.~Long, X.~Zhu, and J.~Jiang, ``Mgae: Marginalized graph
  autoencoder for graph clustering,'' in \emph{Proceedings of the 2017 ACM on
  Conference on Information and Knowledge Management}, 2017, pp. 889--898.

\bibitem{kipf2016variational}
T.~N. Kipf and M.~Welling, ``Variational graph auto-encoders,'' in
  \emph{Advances in Neural Information Processing Systems Workshops}, 2016.

\bibitem{hassani2020contrastive}
K.~Hassani and A.~H. Khasahmadi, ``Contrastive multi-view representation
  learning on graphs,'' in \emph{International Conference on Machine Learning},
  2020, pp. 4116--4126.

\bibitem{tschannen2019mutual}
M.~Tschannen, J.~Djolonga, P.~K. Rubenstein, S.~Gelly, and M.~Lucic, ``On
  mutual information maximization for representation learning,'' in
  \emph{International Conference on Learning Representations}, 2020.

\bibitem{hjelm2018learning}
R.~D. Hjelm, A.~Fedorov, S.~Lavoie-Marchildon, K.~Grewal, P.~Bachman,
  A.~Trischler, and Y.~Bengio, ``Learning deep representations by mutual
  information estimation and maximization,'' in \emph{International Conference
  on Learning Representations}, 2019.

\bibitem{shechtman2007matching}
E.~Shechtman and M.~Irani, ``Matching local self-similarities across images and
  videos,'' in \emph{Proceedings of the IEEE Conference on Computer Vision and
  Pattern Recognition}, 2007, pp. 1--8.

\bibitem{lampert2013attribute}
C.~H. Lampert, H.~Nickisch, and S.~Harmeling, ``Attribute-based classification
  for zero-shot visual object categorization,'' \emph{IEEE Transactions on
  Pattern Analysis and Machine Intelligence}, vol.~36, no.~3, pp. 453--465,
  2013.

\bibitem{yu2017protein}
L.~Yu, Y.~Zhang, I.~Gutman, Y.~Shi, and M.~Dehmer, ``Protein sequence
  comparison based on physicochemical properties and the position-feature
  energy matrix,'' \emph{Scientific Reports}, vol.~7, no.~1, pp. 1--9, 2017.

\bibitem{kim2010walk}
S.~Kim, J.~Yoon, J.~Yang, and S.~Park, ``Walk-weighted subsequence kernels for
  protein-protein interaction extraction,'' \emph{BMC Bioinformatics}, vol.~11,
  no.~1, pp. 1--21, 2010.

\bibitem{chen2019hybrid}
B.~Chen and W.~Deng, ``Hybrid-attention based decoupled metric learning for
  zero-shot image retrieval,'' in \emph{Proceedings of the IEEE Conference on
  Computer Vision and Pattern Recognition}, 2019, pp. 2750--2759.

\bibitem{opitz2018deep}
M.~Opitz, G.~Waltner, H.~Possegger, and H.~Bischof, ``Deep metric learning with
  bier: Boosting independent embeddings robustly,'' \emph{IEEE Transactions on
  Pattern Analysis and Machine Intelligence}, vol.~42, no.~2, pp. 276--290,
  2018.

\bibitem{kim2018attention}
W.~Kim, B.~Goyal, K.~Chawla, J.~Lee, and K.~Kwon, ``Attention-based ensemble
  for deep metric learning,'' in \emph{Proceedings of the European Conference
  on Computer Vision}, 2018, pp. 736--751.

\bibitem{xu2020towards}
X.~Xu, Z.~Wang, C.~Deng, H.~Yuan, and S.~Ji, ``Towards improved and
  interpretable deep metric learning via attentive grouping,'' \emph{arXiv
  preprint arXiv:2011.08877}, 2020.

\bibitem{xu2018powerful}
K.~Xu, W.~Hu, J.~Leskovec, and S.~Jegelka, ``How powerful are graph neural
  networks?'' in \emph{International Conference on Learning Representations},
  2019.

\bibitem{xie2021self}
Y.~Xie, Z.~Xu, J.~Zhang, Z.~Wang, and S.~Ji, ``Self-supervised learning of
  graph neural networks: A unified review,'' \emph{arXiv preprint
  arXiv:2102.10757}, 2021.

\bibitem{nowozin2016f}
S.~Nowozin, B.~Cseke, and R.~Tomioka, ``f-gan: Training generative neural
  samplers using variational divergence minimization,'' in \emph{Advances in
  Neural Information Processing Systems}, 2016.

\bibitem{cheng2020club}
P.~Cheng, W.~Hao, S.~Dai, J.~Liu, Z.~Gan, and L.~Carin, ``Club: A contrastive
  log-ratio upper bound of mutual information,'' in \emph{International
  Conference on Machine Learning}, 2020, pp. 1779--1788.

\bibitem{NEURIPS2019_9015}
A.~Paszke, S.~Gross, F.~Massa, A.~Lerer, J.~Bradbury, G.~Chanan, T.~Killeen,
  Z.~Lin, N.~Gimelshein, L.~Antiga, A.~Desmaison, A.~Kopf, E.~Yang, Z.~DeVito,
  M.~Raison, A.~Tejani, S.~Chilamkurthy, B.~Steiner, L.~Fang, J.~Bai, and
  S.~Chintala, ``Pytorch: An imperative style, high-performance deep learning
  library,'' in \emph{Advances in Neural Information Processing Systems}, 2019,
  pp. 8024--8035.

\bibitem{Fey/Lenssen/2019}
M.~Fey and J.~E. Lenssen, ``Fast graph representation learning with {PyTorch
  Geometric},'' in \emph{International Conference on Learning Representations
  Workshop}, 2019.

\bibitem{grover2016node2vec}
A.~Grover and J.~Leskovec, ``node2vec: Scalable feature learning for
  networks,'' in \emph{Proceedings of the 22nd ACM SIGKDD International
  Conference on Knowledge Discovery and Data Mining}, 2016, pp. 855--864.

\bibitem{adhikari2018sub2vec}
B.~Adhikari, Y.~Zhang, N.~Ramakrishnan, and B.~A. Prakash, ``Sub2vec: Feature
  learning for subgraphs,'' in \emph{Pacific-Asia Conference on Knowledge
  Discovery and Data Mining}.\hskip 1em plus 0.5em minus 0.4em\relax Springer,
  2018, pp. 170--182.

\bibitem{narayanan2017graph2vec}
A.~Narayanan, M.~Chandramohan, R.~Venkatesan, L.~Chen, Y.~Liu, and S.~Jaiswal,
  ``graph2vec: Learning distributed representations of graphs,'' \emph{arXiv
  preprint arXiv:1707.05005}, 2017.

\bibitem{sterling2015zinc}
T.~Sterling and J.~J. Irwin, ``Zinc 15--ligand discovery for everyone,''
  \emph{Journal of Chemical Information and Modeling}, vol.~55, no.~11, pp.
  2324--2337, 2015.

\bibitem{wu2018moleculenet}
Z.~Wu, B.~Ramsundar, E.~N. Feinberg, J.~Gomes, C.~Geniesse, A.~S. Pappu,
  K.~Leswing, and V.~Pande, ``Moleculenet: a benchmark for molecular machine
  learning,'' \emph{Chemical Science}, vol.~9, no.~2, pp. 513--530, 2018.

\bibitem{hu2019strategies}
W.~Hu, B.~Liu, J.~Gomes, M.~Zitnik, P.~Liang, V.~Pande, and J.~Leskovec,
  ``Strategies for pre-training graph neural networks,'' in \emph{International
  Conference on Learning Representations}, 2020.

\end{thebibliography}
	\bibliographystyle{IEEEtran}
\end{document}